\documentclass[runningheads]{llncs}
\usepackage{graphicx}

\usepackage{tikz}
\usepackage{comment}
\usepackage{amsmath,amssymb} 
\usepackage{color}

\usepackage[numbers]{natbib}

\usepackage[accsupp]{axessibility}  

\usepackage{hyperref}
\usepackage[capitalise]{cleveref}
\usepackage{bm}
\usepackage{multirow}
\usepackage{caption}
\usepackage{subcaption}
\usepackage{booktabs}
\usepackage{url}
\usepackage{wrapfig}

\newcommand{\etal}{\textit{et al.}}
\newcommand{\eg}{\textit{e.g.}}

\makeatletter
\def\@fnsymbol#1{\ensuremath{\ifcase#1 \or * \or \dagger \or \ddagger \else\@ctrerr\fi}}
\makeatother

\begin{document}
\pagestyle{headings}
\mainmatter
\def\ECCVSubNumber{2902}  

\title{UNIF: United Neural Implicit Functions for Clothed Human Reconstruction and Animation}


	\titlerunning{UNIF}
	%
	\author{
		Shenhan Qian\inst{1,2}\thanks{Work conducted during an internship at ZMO AI Inc.} \quad
		Jiale Xu\inst{2} \quad \\
		Ziwei Liu\inst{3} \quad
		Liqian Ma\inst{1}\thanks{Corresponding author.}  \quad
		Shenghua Gao\inst{2,4,5}\\
		{\tt\small 
		\{qianshh, xujl1, gaoshh\}@shanghaitech.edu.cn \\
		zwliu.hust@gmail.com \quad 
		liqianma.scholar@outlook.com}
	}
	\authorrunning{S. Qian et al.}
	%
	\institute{
		ZMO AI Inc. \and
		ShanghaiTech University \and
		S-Lab, Nanyang Technological University \and
		Shanghai Engineering Research Center of Intelligent Vision and Imaging \and
		Shanghai Engineering Research Center of Energy Efficient and Custom AI IC
	}
\maketitle

\begin{abstract}
	We propose united implicit functions (UNIF), a part-based method for clothed human reconstruction and animation with raw scans and skeletons as the input. Previous part-based methods for human reconstruction rely on ground-truth part labels from SMPL and thus are limited to minimal-clothed humans. In contrast, our method learns to separate parts from body motions instead of part supervision, thus can be extended to clothed humans and other articulated objects. Our Partition-from-Motion is achieved by a bone-centered initialization, a bone limit loss, and a section normal loss that ensure stable part division even when the training poses are limited. We also present a minimal perimeter loss for SDF to suppress extra surfaces and part overlapping. Another core of our method is an adjacent part seaming algorithm that produces non-rigid deformations to maintain the connection between parts which significantly relieves the part-based artifacts. Under this algorithm, we further propose ``Competing Parts'', a method that defines blending weights by the relative position of a point to bones instead of the absolute position, avoiding the generalization problem of neural implicit functions with inverse LBS (linear blend skinning). We demonstrate the effectiveness of our method by clothed human body reconstruction and animation on the CAPE and the ClothSeq datasets. Our code is available at \url{https://github.com/ShenhanQian/UNIF.git}.
	
\end{abstract}


\begin{figure}
	\centering
	\includegraphics[width=0.9\linewidth]{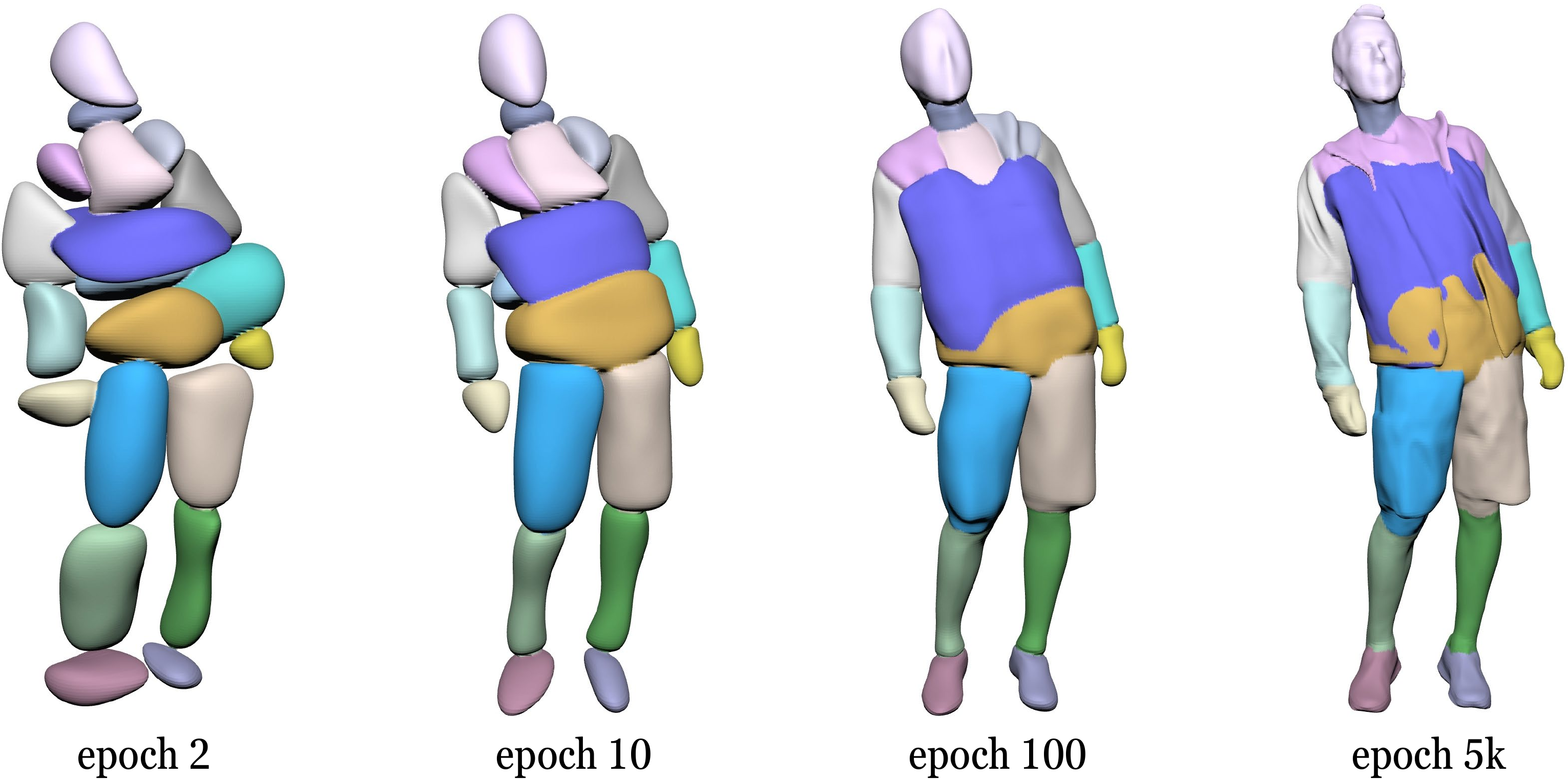}
	\caption{The evolution of the learned parts of our model.}
	\label{fig:evolution-of-parts}
\end{figure}

\section{Introduction}
As residents of the 21st century, we are embracing a new life in the virtual world, digitizing everything around us.
Recent research interest in human body reconstruction and animation increases dramatically.
A popular human body model is SMPL~\cite{loper2015smpl}, which models minimal-clothed human bodies across genders and figures. Later methods extend SMPL~\cite{loper2015smpl} to the clothed human body by overlying vertex offsets~\cite{ma2020learning, liu2019liquid} or attaching template clothing meshes~\cite{patel2020tailornet}. However, for complex clothes, the fixed topology of SMPL~\cite{loper2015smpl} mesh and the predefined template clothing limit the expressiveness. Recently, the rise of neural implicit representations~\cite{chen2019learning, mescheder2019occupancy, park2019deepsdf} indicates a higher modeling fidelity and flexibility. These models take in a point position and output an indicator of the geometry such as occupancy and SDF (signed distance function), theoretically supporting an infinitely high resolution. The infinity of resolution is perfect for fidelity but a disaster for skinning since we can no longer store the LBS (linear blend skinning) weights for every point. Although recent methods use another neural implicit function to learn the weights~\cite{saito2021scanimate, mihajlovic2021leap, peng2021animatable, tiwari2021neural}, they generalize poorly under unseen poses because the LBS weights of a point vary along with the pose.

Besides learning a whole shape and deforming it with LBS, we can also model an object with separate parts. NASA~\cite{deng2020nasa} models the human body with several occupancy networks, each of which is bound to a joint. Therefore, when the skeleton moves, the learned shape is articulated. However, a key limitation of NASA~\cite{deng2020nasa} and later part-based methods~\cite{alldieck2021imghum, lombardi2021latenthuman} is that they rely on SMPL's LBS weights for part division, therefore still limited to minimal-clothed human reconstruction. Another shortage of previous part-based methods is that they model the non-rigid deformation crudely by simply feeding the positions of posed joints or similar pose descriptors into the networks. This results in an overfitted model that produces artifacts under novel poses especially when the training poses are limited. 

To push the boundary of part-based methods, we propose UNIF (united neural implicit functions), a method that learns the shape of an object with multiple neural implicit functions. Our method features two novelties: 1) UNIF learns to decouple parts from a whole shape with no need for ground-truth partition labels; 2) UNIF models non-rigid deformations by considering the interaction between parts.

To illustrate the basic idea of automatic part division, let us consider an arm moving relative to the body. For a part-based method, we expect the arm and the body to be modeled by two separate networks. In case they are captured by one network, they will always move as a rigid one, then the model will not be able to reconstruct the same shape when the arm moves. Therefore, when minimizing the surface reconstruction loss with changing poses, we are pushing the networks to converge into separate rigid parts. We call this process \textit{Partition-from-Motion}. However, when the training poses are limited, \textit{e.g.}, the subject in the raw scans never moves the arms, then there will be no driving force to decouple the arm from the body. This is not a big issue for reconstruction but unacceptable for novel-pose animation since the arms can never move. To ensure a good body partition when the training poses are limited, we propose a bone limit loss and a section normal loss that constrain the boundary and the normal of each part by its neighboring joints. These terms significantly enhance the stability of our Partition-from-Motion. Furthermore, motivated by PHASE \cite{lipman2021phase}, we derive a minimal perimeter loss on SDF to suppress extra parts and hidden surfaces, which also contributes to a high-quality reconstruction.

Partition-from-Motion helps us separate rigid parts, but this is insufficient because non-rigid deformations are not negligible for human bodies and clothes. We propose an APS (adjacent part seaming) algorithm that deforms points to maintain the connection between parts. APS greatly relieves artifacts such as cracks and exposure of hidden surfaces. Alike other non-rigid deformation algorithms, APS also needs to define the blending weights of a point. Differently, we define blending weights not by the absolute position but by the relative position of a point to each bone and the competition between bones. Such a local definition of blending weights avoids overfitting of absolute positions, generalizing better to unseen poses.

Overall, our contributions can be summarized as:
\begin{itemize}
	\item We propose united neural implicit functions (UNIF) for clothed human reconstruction and animation from raw scan sequences.
	\item We decouple rigid parts without partition labels and enhance the robustness with carefully designed initialization and regularization strategies.
	\item We design an adjacent part seaming (APS) algorithm for non-rigid deformation based on a localized definition of blending weights (Competing Parts).
	\item We show the effectiveness of our method by clothed human reconstruction and animation on the CAPE~\cite{ma2020learning} and the ClothSeq~\cite{tiwari2021neural} dataset.
\end{itemize}

\section{Related Work}
Our method adopts compound neural implicit functions for human body reconstruction and animation with special attention to part division and non-rigid deformation.
\subsection{Neural Implicit Functions}
Compared to classic geometry representations such as meshes, point clouds, and voxels that are stored as discrete elements, neural implicit functions \cite{mescheder2019occupancy, park2019deepsdf, chen2019learning, atzmon2020sal, gropp2020igr, mildenhall2020nerf} are stored with neural networks. They take in the coordinate of a point and output an indicator of geometry, appearance, or other properties. Early methods need dense supervision of occupancy or SDF \cite{mescheder2019occupancy, park2019deepsdf, chen2019learning}. Later works make it possible to learn smooth surfaces with sparse supervision~\cite{atzmon2020sal, gropp2020igr, lipman2021phase}. SAL~\cite{atzmon2020sal} proposes a geometric initialization to realize signed distance learning with unsigned ground-truth data. Benefitting from the Eikonal loss to maintain a valid SDF field, IGR~\cite{gropp2020igr} only takes raw scans or triangle soups as the input. Lipman~\etal~\cite{lipman2021phase} unifies SDF and occupancy and proposes a minimal perimeter loss to encourage tight surfaces. Our method follows this line of methods for its lower requirement for the data. It is also possible to model a scene or an object from 2D images without explicitly decoupling the geometry, appearance, and lighting condition \cite{mildenhall2020nerf, yariv2020multiview, oechsle2021unisurf, yariv2021volume, wang2021neus}. For the usage of compound implicit functions, existing trials mainly lie in template-based shape learning~\cite{genova2019learning, genova2020local}.

\subsection{Human Body Reconstruction and Animation}
As the most popular mesh-based human body model, SMPL~\cite{loper2015smpl} and its variations \cite{li2017learning, romero2017mano, pavlakos2019expressive} dominate the area of human body reconstruction for its expressiveness and flexibility, supporting innumerable downstream task~\cite{liu2019liquid, patel2020tailornet, atzmon2020sal, gropp2020igr, peng2021neural, peng2021animatable, liu2021neural}. Since the new trend of neural implicit functions for shape learning, several papers \cite{deng2020nasa, mihajlovic2021leap, alldieck2021imghum, lombardi2021latenthuman, chen2021snarf} have attempted to substitute SMPL with an implicit counterpart for higher fidelity and flexibility. Besides the minimal-clothed human body, later works also use neural implicit functions to model clothed humans \cite{saito2021scanimate, tiwari2021neural, palafox2021npms, peng2021neural, peng2021animatable}. 

For body animation, there exist two types of pose representation - latent vector and skeleton. SAL \cite{atzmon2020sal}, IGR \cite{gropp2020igr}, and NPMs \cite{palafox2021npms} model body poses with a latent space, which is especially useful when no skeleton is available. But they only support interpolation between poses instead of direct animation. As to pose interpolation in the latent space, Atzmon~\etal~\cite{atzmon2021augmenting} regularize the deformation field concerning the latent vector to maintain the as-rigid-as-possible property.

Among the skeleton-based methods, the mainstream practice is to learn a canonical shape and animate it with LBS (linear blend skinning). However, since a neural implicit function lacks point-wise correspondences, a forward and a backward skinning network are introduced \cite{saito2021scanimate, mihajlovic2021leap, peng2021animatable, tiwari2021neural} to save LBS weights for the bidirectional mapping between a posed shape and the canonical shape. The main limitation here is the poor generalization ability of inverse LBS since the LBS weights vary when the pose changes. SCANimate \cite{saito2021scanimate} and LEAP \cite{mihajlovic2021leap} use the cycle consistency to regularize the learned neural skinning weights. In contrast, SNARF~\cite{chen2021snarf} only learns the stable forward skinning weights and solves backward skinning by iteratively minimizing the cycle consistency error.

\begin{figure}[t]
	\centering
	\includegraphics[width=1.0\linewidth]{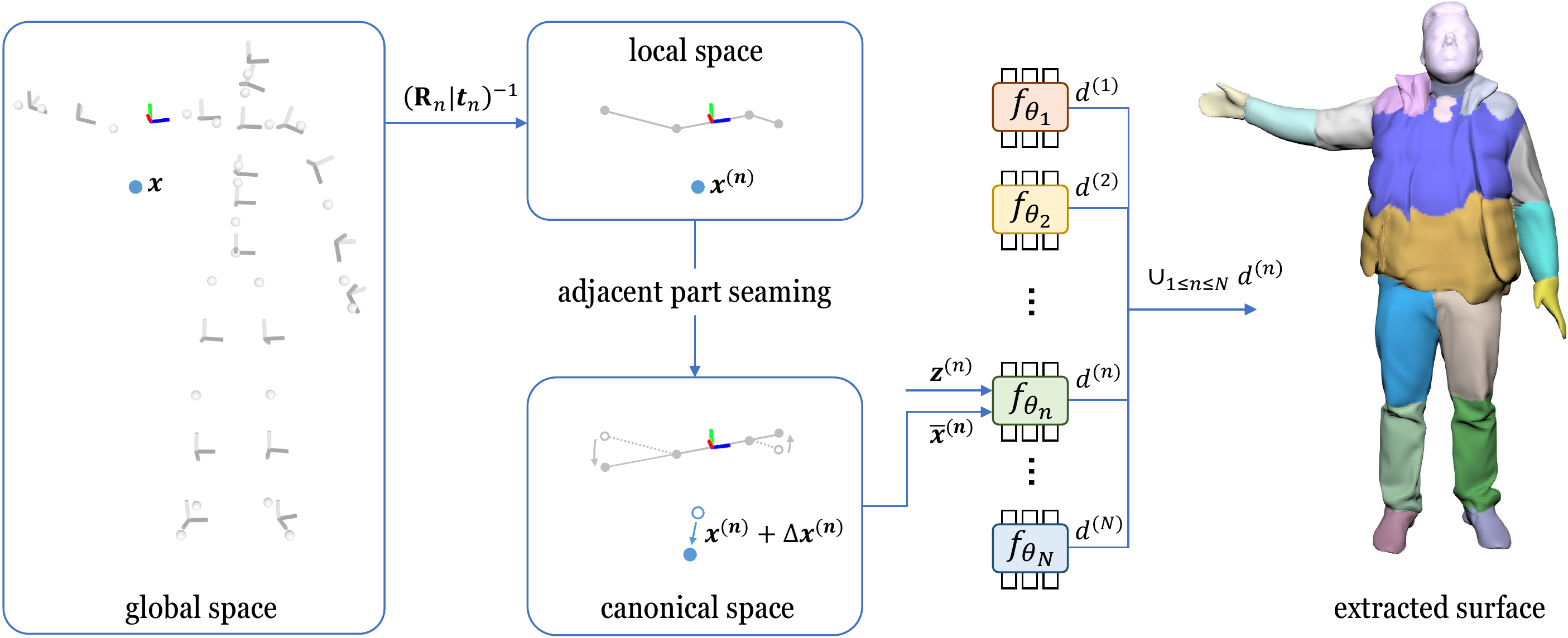}
	\caption{The pipeline of our method. Given a point $\bm{x}$, we first transform it into the local space of each bone and then apply our adjacent part seaming algorithm to obtain its position $\bar{\bm{x}}^{(n)}$ in the canonical space  of the $n$-th bone. Each neural implicit function takes the position $\bar{\bm{x}}^{(n)}$ and a pose condition vector $\bm{z}^{(n)}$ to predict the SDF value of a part $d^{(n)}$. The final output of our method is the union of all output.}
	\label{fig:pipeline}
\end{figure}

Aside from LBS-based methods, another series of methods model the human body with separate parts. NASA~\cite{deng2020nasa} merges the output of a group of occupancy networks, each anchored on a joint of the body. Both LatentHuman \cite{lombardi2021latenthuman} and imGHUM~\cite{alldieck2021imghum} train combinational signed distance functions on muti-subject data. LatentHuman~\cite{lombardi2021latenthuman} pays special attention to relieve part based artifacts, while imGHUM~\cite{alldieck2021imghum} provides additional controllability on hands and expressions. A common feature of the above part-based methods is that they all rely on the LBS weights of SMPL to partite the body. In contrast, we learn part division from body motion. As to the non-rigid deformation, previous part-based methods \cite{deng2020nasa, lombardi2021latenthuman} simply feed skeleton states as an input of networks, leading to limited pose generalization ability.

\section{United Neural Implicit Functions}

The input of our method is a sequence of point clouds, which captures the shapes of a person in varying poses. For each frame, we fit the body skeleton (\eg, the skeleton of SMPL \cite{loper2015smpl}) represented by the orientations and translations of body joints. Then, we set up local coordinate systems based on the skeleton and learn a neural implicit function in each local space.

We illustrate the pipeline of our method in \cref{fig:pipeline}. For a point $\bm{x}$ in the global space, we first transform it to the local space of each bone and get $\bm{x}^{(n)}$. Then we deform the point by an offset $\Delta\bm{x}^{(n)}$ with an adjacent part seaming (APS) algorithm and get its position $\bar{\bm{x}}^{(n)}$ in the canonical space of the $n$-th bone. Finally, we feed the position $\bar{\bm{x}}^{(n)}$ and a pose condition vector $\bm{z}^{(n)}$ to each neural implicit function and take the union of their output.

\subsection{Shape Representation and Learning}
\label{sec:shape-representatioin-and-learning}
Our united neural implicit functions are based on IGR \cite{gropp2020igr}, which adopts a single neural network to model the surface of an object. Given a point cloud $\mathcal{X}=\left\{\bm{x}_{i}\right\}_{i \in I} \subset \mathbb{R}^{3}$ and corresponding surface normals $\mathcal{N}=\left\{\bm{n}_{i}\right\}_{i \in I} \subset \mathbb{R}^{3}$, IGR \cite{gropp2020igr} optimizes the parameters $\theta$ of an MLP $f_{\theta}(\bm{x})$ to approximate the signed distance function of the surface behind the point cloud $\mathcal{X}$ with the loss
\begin{equation}
	\mathcal{L}=\mathcal{L}_\text{recon} + \lambda_\text{unit}\mathcal{L}_\text{unit},
	\label{eq:igr}
\end{equation}
where
\begin{equation}
	\mathcal{L}_\text{recon}=\frac{1}{|I|} \sum_{i \in I}\left(\left|f_{\theta}\left(\boldsymbol{x}_{i}\right)\right|+\lambda_\text{normal}\left\|\nabla_{\boldsymbol{x}} f_{\theta}\left(\boldsymbol{x}_{i}\right)-\boldsymbol{n}_{i}\right\|_{2}\right),
\end{equation}
\begin{equation}
	\mathcal{L}_\text{unit}= \mathbb{E}_{\boldsymbol{x}}\left(\left\|\nabla_{\boldsymbol{x}} f_{\theta}(\boldsymbol{x})\right\|_{2} -1\right)^{2}.
\end{equation}
$\mathcal{L}_\text{recon}$ supervises the zero-level set of $f$ to go across $\mathcal{X}$ with the given normals $\mathcal{N}$. $\mathcal{L}_\text{unit}$ encourages the gradient of $f$ to be unit-norm, which is necessary for a signed distance function.

For our UNIF model, we use $N$ ($N=20$) separate MLPs ($f_{\theta_1}, \dots, f_{\theta_N}$), each learns the SDF of a body part. Given a point $\bm{x}$ from the input point cloud $\mathcal{X}$, the output of UNIF is
\begin{equation}
	d = \cup_{1\leq n \leq N} d^{(n)}, \quad\text{with } d^{(n)} = f_{\theta_n}\left(\bm{x}^{(n)} \right).
\end{equation}
$\cup$ is an union operation on the output of all networks. Geometrically, the union of multiple signed distance functions is the minimum of all:
\begin{equation}
	d = \min_{1\leq n \leq N}{d^{(n)}}.
\end{equation}
To ease learning and enhance robustness, we use an improved union operation, which is presented in the supplementary material.
$\bm{x}^{(n)}$ is the local point position for the $n$-th part with
\begin{equation}
	\bm{x}^{(n)} = \mathbf{R}_n^{T}(\bm{x} - \bm{t}_n),
\end{equation}
where $\mathbf{R}_n$ and $\bm{t}_n$ are the global orientation and translation of the $n$-th coordinate system. 

Finally, the supervision on our UNIF model becomes
\begin{equation}
	\mathcal{L}=\mathcal{L}_\text{recon} + \lambda_\text{unit}\mathcal{L}_\text{unit},
	\label{eq:unif-loss}
\end{equation}
where
\begin{equation}
	\mathcal{L}_\text{recon}=\frac{1}{|I|} \sum_{i \in I}\left(\left|d\right|+\lambda_\text{normal}\left\|\nabla_{\boldsymbol{x}} d-\boldsymbol{n}_{i}\right\|_{2}\right) \quad \left( \lambda_{\text{normal}}=0.01\right),
\end{equation}
\begin{equation}
	\mathcal{L}_\text{unit}= \mathbb{E}_{\boldsymbol{x}}\left(\left\|\nabla_{\boldsymbol{x}} d\right\|_{2}-1\right)^{2} + \frac{1}{N} \sum_{n=1}^{N} \mathbb{E}_{\boldsymbol{x}}\left(\left\|\nabla_{\boldsymbol{x}} d^{(n)}\right\|_{2}-1\right)^{2}.
\end{equation}
The unit-gradient-norm loss $\mathcal{L}_\text{unit}$ has two terms. The first term is applied on the SDF after the union operation ($d$), and the second term is applied on the output of each part ($d^{(n)}$). Both are necessary according to our experiments.

\subsection{Partition-from-Motion}
\label{sec:partition-from-motion}

Unlike previous methods \cite{deng2020nasa, alldieck2021imghum, lombardi2021latenthuman} that use ground-truth partition labels from SMPL \cite{loper2015smpl}, we exploit separating parts automatically while learning the entire shape. The key to achieving this is using SDF instead of occupancy because occupancy is constantly zero for locations away from the surface, while SDF provides distance information so that we can determine which part is closer to the query point and then optimize that part to go across the point. This reveals an implicit hypothesis of our method: a point should be assigned to the closest part to it. However, the SDF of a part is randomly initialized and thus may not provide correct distance information at the beginning of training. Therefore, we propose a bone-centered initialization.

\begin{figure}[t]
	\centering
	\begin{subfigure}[t]{0.43\textwidth}
		\centering
		\includegraphics[width=0.63\textwidth]{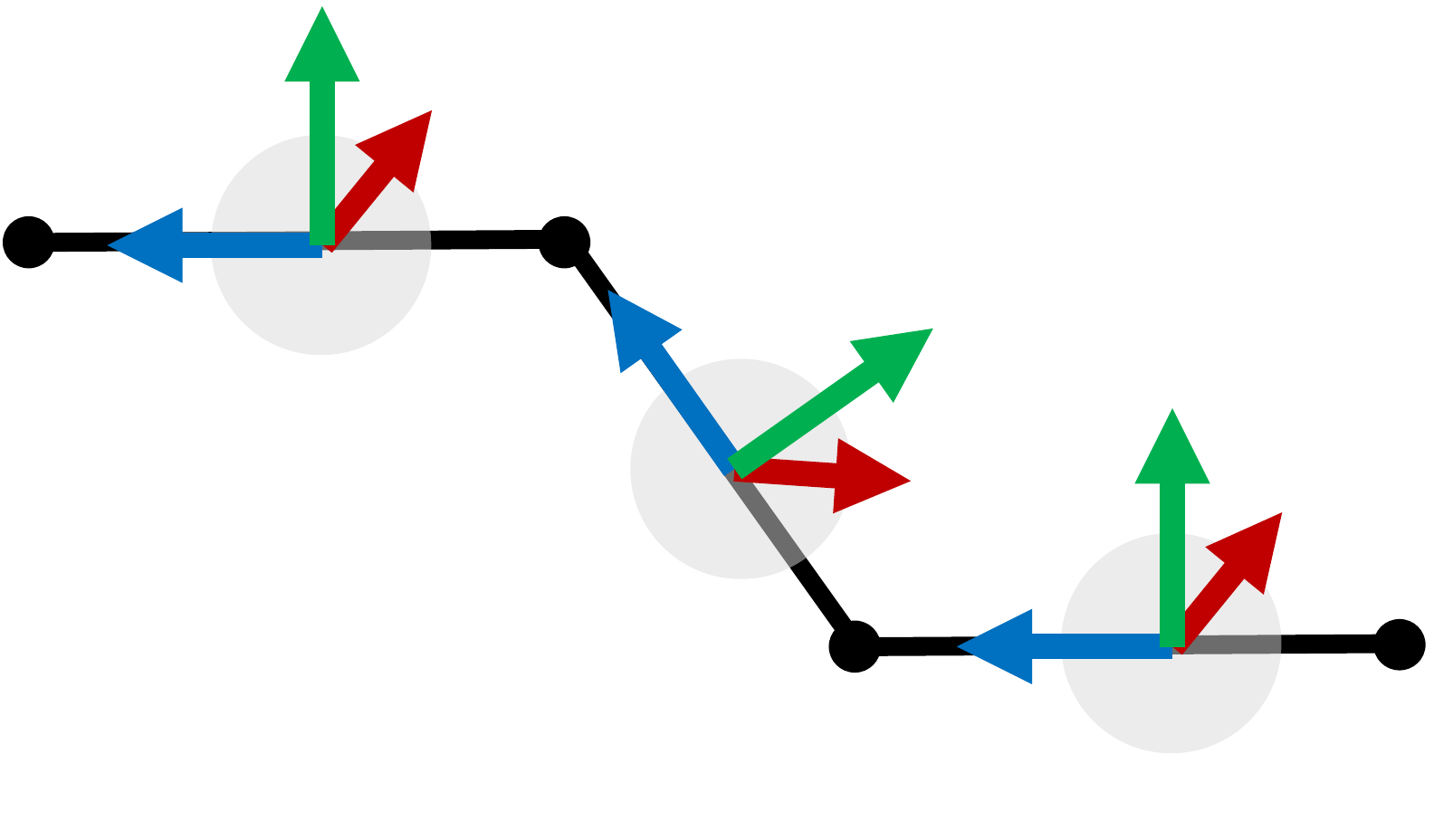}
		\caption{Bone-centered initialization. Each neural implicit function is initialized to a small sphere at the center of a bone.}
		\label{fig:bone-centered-init}
	\end{subfigure}
	\hfill
	\begin{subfigure}[t]{0.51\textwidth}
		\centering
		\includegraphics[width=0.55\textwidth]{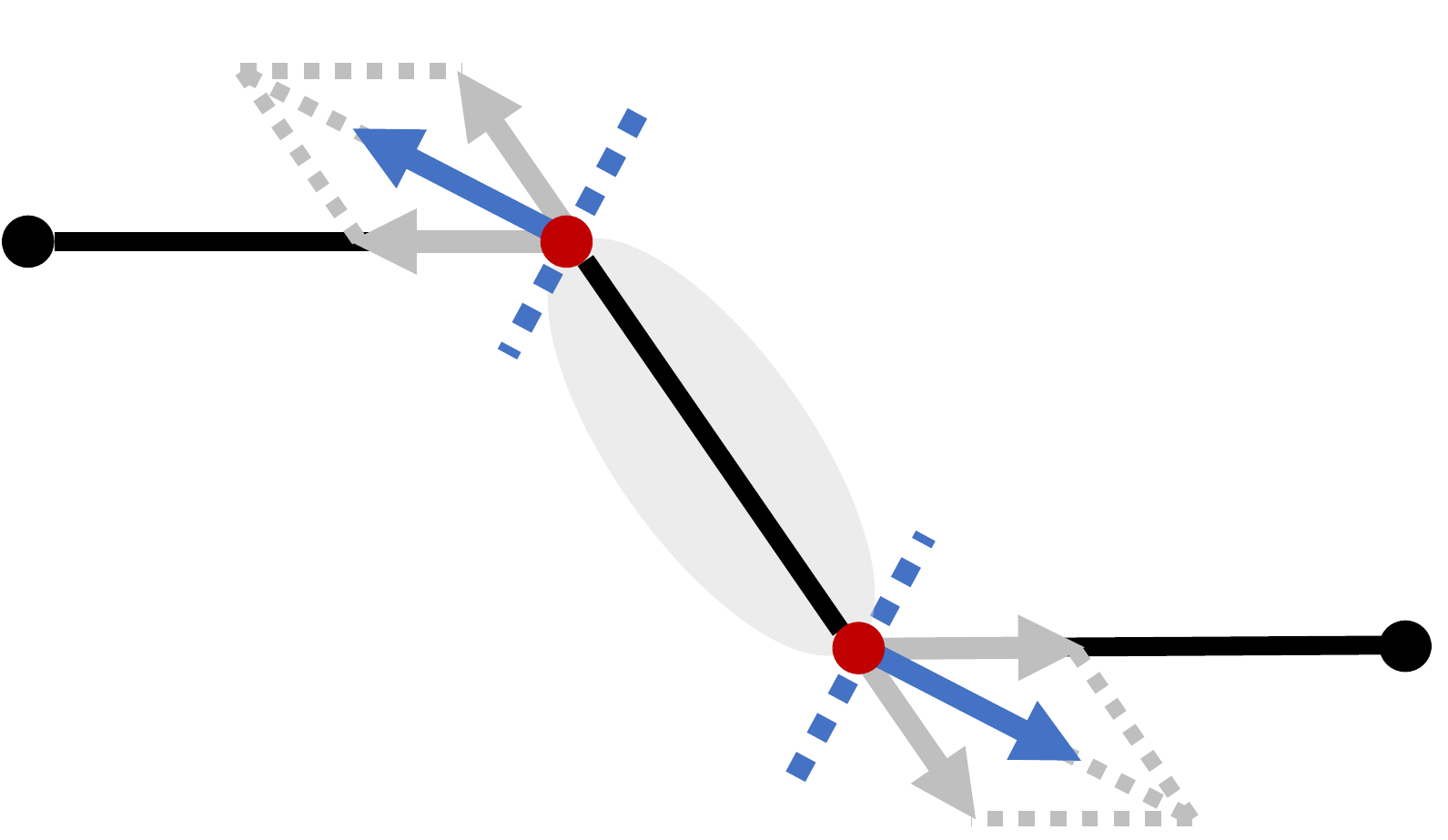}
		\caption{Bone limit loss and section normal loss. The boundary (in red) and section normals (in blue) of a part are constrained by its neighboring joints.}
		\label{fig:bone-limit-and-section-normal}
	\end{subfigure}
	\caption{Improving Partition-from-Motion with skeleton-based priors.}
	\label{fig:partition-from-motion}
\end{figure}

\subsubsection{Bone-centered initialization.}
\label{sec:bone-centered-initialization}
We set up local coordinate systems at the center of bones and use the geometric initialization \cite{atzmon2020sal} to turn each part into a small sphere ($r=0.01$) at the bone center (\cref{fig:bone-centered-init}). Then, parts are not intersected, and the SDF of a part approximately equals the distance to the bone center. This ensures that most points are assigned to the right part when training begins.

\subsubsection{Bone limit loss and section normal loss.}
With a proper initialization, we can already separate parts, but the quality and stability of body partition highly depend on the variance of training poses. For example, when two parts barely have relative motions in the training set, they are at high risk of overlapping. This leads to artifacts when the model is animated under novel poses. Therefore, we propose a bone limit loss
\begin{equation}
	\mathcal{L}_\text{lim} =\frac{1}{N \cdot \left|J^{(n)}\right|} \sum_{n=1}^{N}\sum_{j \in {J^{(n)}}}\left|d_j^{(n)}\right|,
\end{equation}
and a section normal loss
\begin{equation}
	\mathcal{L}_\text{sec} =\frac{1}{N \cdot \left|J^{(n)}\right|} \sum_{n=1}^{N}\sum_{j \in {J^{(n)}}} \left\|\nabla_{\boldsymbol{x}} d_j^{(n)} -\boldsymbol{n}_j^{(n)}  \right\|_{2},
\end{equation}
where $J^{(n)}$ is the $n$-th bone's adjacent joints and $\left|J^{(n)}\right|$ is the number of its adjacent joints; $d_j^{(n)}$ is the predicted SDF at joint $j$; $\bm{n}_j^{(n)}$ is the section normal at joint $j$ derived from the angle between adjacent bones. As illustrated by \cref{fig:bone-limit-and-section-normal}, these two terms utilize the positions of joints as a prior to limit the range of a part along the axis of its bone and the normal of the sections.

\subsubsection{Minimal perimeter loss.}
In experiments, our method often produces artifacts like extra surfaces, which are due to the insufficiency of the IGR \cite{gropp2020igr} loss. 
Considering \cref{eq:igr}, the reconstruction term $\mathcal{L}_\text{recon}$ ensures a zero value at the positions of raw scans and the unit-norm term $\mathcal{L}_\text{unit}$ regularize the gradient of the neural field, but neither punish extra surfaces where no scan points lie. Inspired by PHASE \cite{lipman2021phase}, we propose a minimal perimeter loss specifically for SDF:
\begin{equation}
	\mathcal{L}_{\text{perim}} = \mathbb{E}_{\boldsymbol{x}}\left\|\nabla_{\boldsymbol{x}} \sigma(d)\right\|^{2} + \frac{1}{N} \sum_{n=1}^{N} \mathbb{E}_{\boldsymbol{x}}\left\|\nabla_{\boldsymbol{x}} \sigma(d^{(n)})\right\|^{2},
\end{equation}
where $\sigma(x) = \frac{1}{1 + e^{- \beta x}}$ (we use $\beta=10$). This minimal perimeter loss $\mathcal{L}_{\text{perim}}$ is applied both globally and locally, similar to $\mathcal{L}_{\text{unit}}$. The global term ensures the tightness of the overall shape, while the local term suppresses the extra surfaces hidden behind the overall shape. We leave further discussion of this loss in the supplementary material.

\subsection{Adjacent Part Seaming}
\label{sec:adjacent_parts_seaming}

Now, we are able to learn separate parts automatically with proper initialization and regularization. But be aware that the entire model is moving as rigid parts, obviously insufficient for either human bodies or clothes. To support non-rigid deformation, previous part-based methods \cite{deng2020nasa, alldieck2021imghum, lombardi2021latenthuman} feed a descriptor of joints into networks. We construct a similar descriptor by first transforming the orientation matrices and translation vectors of all joints into each part's local space, then flattening and concatenating them into a pose condition vector
\begin{equation}
	\bm{z}^{(n)} = \oplus_{1\leq j \leq N}{\left( \mathbf{R}_n^T \mathbf{R}_j \oplus \mathbf{R}_n^T \left(\bm{t}_j-\bm{t}_n\right) \right)},
\end{equation}
where $\mathbf{R}_n$ and $\bm{t}_n$ are the orientation and translation of the $n$-th bone; $\mathbf{R}_j$ and $\bm{t}_j$ are the orientation and translation of the $j$-th joint; $N$ is the number of parts and $\oplus$ refers to vector concatenation.
Then, our neural implicit functions become
\begin{equation}
	d^{(n)} = f_{\theta_n}\left( \bm{x}^{(n)}, \bm{z}^{(n)} \right), 1 \leq n \leq N.
\end{equation}
The above pose descriptor does help the network fit the training sequence but generalizes poorly to unseen poses. As demonstrated in recent comparisons \cite{chen2021snarf, tiwari2021neural}, part-based models always produce broken parts in unseen poses.

Then, can we make non-rigid deformations of part-based models generalizable to unseen poses? Here is an observation: when two linked parts have a relative rotation (\cref{fig:adjacent-part-seaming_a}), some regions are squeezed while others are stretched. We believe that explicitly modeling the phenomenon is the key to relieving part-based artifacts. Therefore, we propose the adjacent part seaming (APS) algorithm.

\subsubsection{Adjacent part seaming by local rotations.}

\begin{figure}[t]
	\centering
	\begin{subfigure}[t]{0.28\textwidth}
		\centering
		\includegraphics[width=\textwidth]{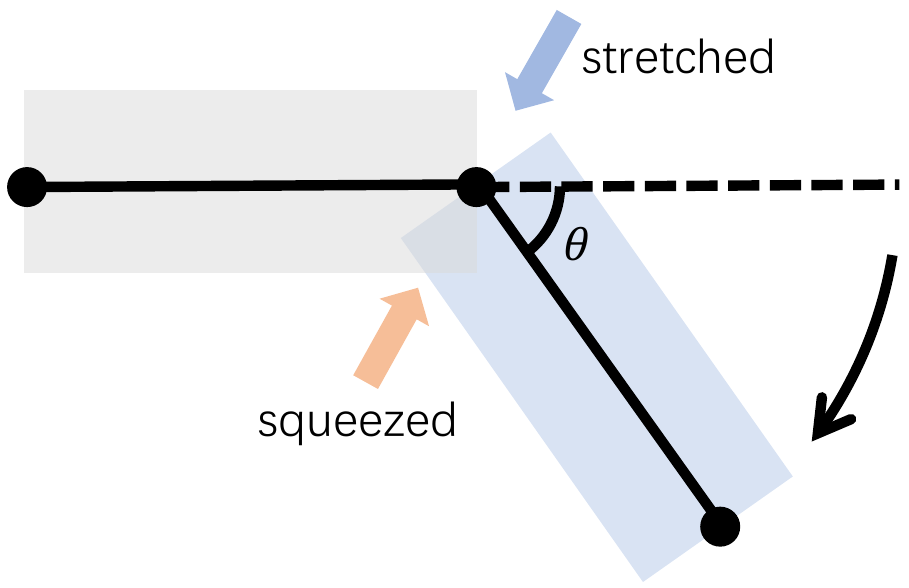}
		\caption{Relative rotation of two linked parts.}
		\label{fig:adjacent-part-seaming_a}
	\end{subfigure}
	\hfill
	\begin{subfigure}[t]{0.28\textwidth}
		\centering
		\includegraphics[width=\textwidth]{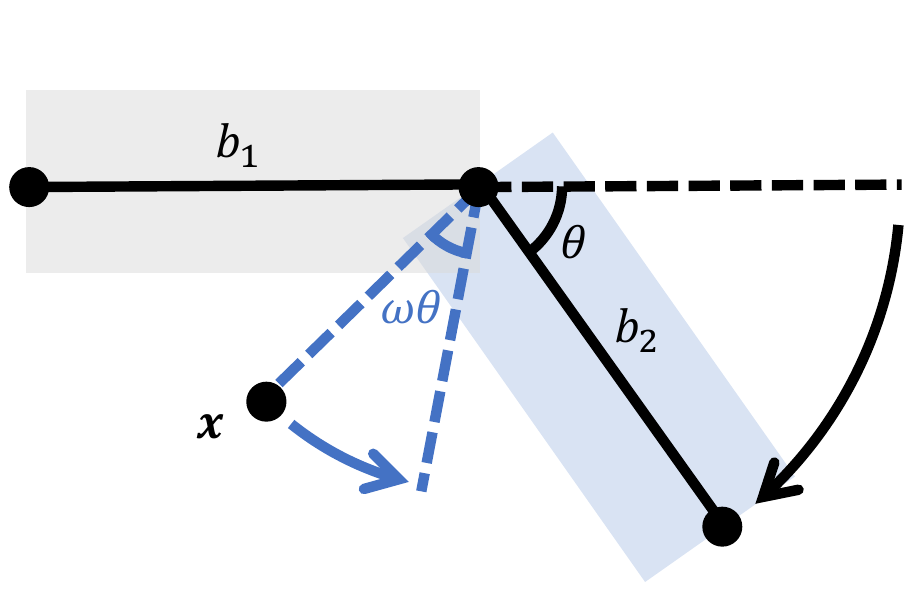}
		\caption{Part seaming for two bones.}
		\label{fig:adjacent-part-seaming_b}
	\end{subfigure}
	\hfill
	\begin{subfigure}[t]{0.40\textwidth}
		\centering
		\includegraphics[width=\textwidth]{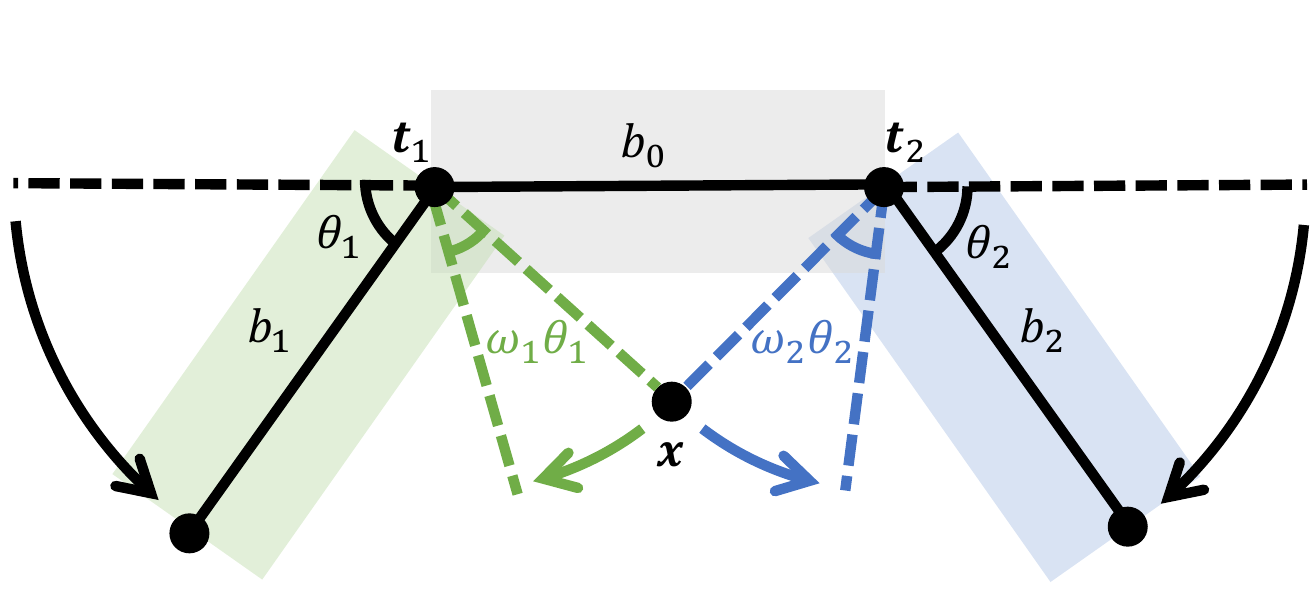}
		\caption{Part seaming for three and more bones}
		\label{fig:adjacent-part-seaming_c}
	\end{subfigure}
	\caption{2D examples to illustrate the process of part seaming.}
\end{figure}

Considering a point $\bm{x}$ on the bone $b_1$ (\cref{fig:adjacent-part-seaming_b}), the adjacent bone $b_2$ has rotated for an angle $\theta$ from the rest pose. What we pursue is the original position of $\bm{x}$ in the rest pose. If the bone $b_1$ is infinitely rigid, then the point $\bm{x}$ on $b_1$ would not have moved no matter the rotation angle of $b_2$. Otherwise, $\bm{x}$ should have rotated for an angle of $w\theta$, where $0 < w < 1$ (we assume the blending weight $w$ known at the moment and will discuss it later). Then, we can obtain the original position of $\bm{x}$ under the rest pose by 
\begin{equation}
	\bar{\bm{x}} = \mathbf{R}_{w \theta}^T \bm{x},
\end{equation}
where $\mathbf{R}_{w\theta}$ is the corresponding rotation matrix for $w\theta$. While we use a 2D example for illustration, the same process can be directly generalized to 3D cases with axis-angles.

For the skeleton of SMPL \cite{loper2015smpl}, a bone can be connected to up to four neighbors. We then consider the case of three connected bones, which also applies to more connections. As shown in \cref{fig:adjacent-part-seaming_c}, when trying to recover a point $\bm{x}$ on the bone $b_0$ to its original position, the point $\bm{x}$ is expected to go through two different rotations. Since the axes of the two rotations are not the same, we cannot simply blend the angles. Instead, we blend the offset vectors:
\begin{equation}
	\Delta\bm{x} = \left(\mathbf{R}_{w_1 \theta_1}^T (\bm{x}-\bm{t}_1) + \bm{t}_1 - \bm{x}\right) + \left(\mathbf{R}_{w_2 \theta_2}^T (\bm{x}-\bm{t}_2) + \bm{t}_2 - \bm{x}\right),
\end{equation}
where $\bm{t}_1$ and $\bm{t}_2$ are the center points of the two rotations.
Then we obtain the original position of $\bm{x}$ by
\begin{equation}
	\bar{\bm{x}} = \bm{x} + \Delta\bm{x}.
\end{equation}
Finally, our neural implicit functions are formulated as
\begin{equation}
	d^{(n)} = f_{\theta_n}\left( \bar{\bm{x}}^{(n)}, \bm{z}^{(n)} \right), 1 \leq n \leq N,
\end{equation}
with
\begin{equation}
	\bar{\bm{x}}^{(n)} = \bm{x}^{(n)} + \Delta\bm{x}^{(n)} = \bm{x}^{(n)} + \sum_{b \in B^{(n)}}{\left(\mathbf{R}_{w_b \theta_b}^T (\bm{x}^{(n)}-\bm{t}_b) + \bm{t}_b - \bm{x}^{(n)}\right)},
\end{equation}
where $B^{(n)}$ is the indices of joints connected to the $n$-th part.

Taking a step back, you may feel the above APS algorithm is quite like inverse LBS since it cancels deformations by reversing transformations as well. But there is a contradiction for inverse LBS: it needs the LBS weights defined in the canonical space before reverting the deformation; but if we already know where to take the LBS weights in the canonical space, we do not need this inverse deformation. To evade this problem, we should avoid saving blending weights by the absolute position. Therefore, we present ``Competing Parts'', a method that defines the blending weights of a point by its relative position to bones so that the blending weights can generalize to arbitrary poses. 

\subsubsection{Blending weights from ``Competing Parts''.}

\begin{wrapfigure}{r}{0.3\textwidth}
	\centering
	\includegraphics[width=0.25\textwidth]{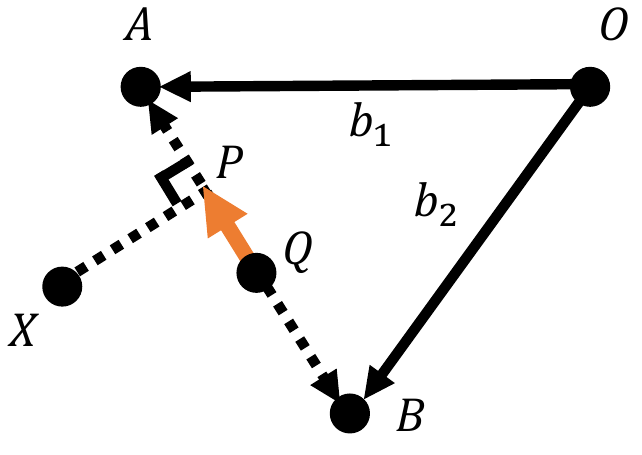}
	\caption{A 2D example to illustrate the definition of rigidness.}
	\label{fig:competing-bones}
\end{wrapfigure}

The basic idea here is that the deformation of a point on a part is the result of the interaction between this part and its adjacent parts. We define the tendency of a point to stay static on the part as the rigidness at this point. Then, we can construct a rigidness field for a part with respect to each of its adjacent part. Taking \cref{fig:competing-bones} as an example, we connect the end points of both bones and split the connecting line with point $Q$ by the ratio of bone lengths, then the rigidness of bone $b_1$ and bone $b_2$ at the point $X$ are defined as

\begin{equation}
	r_1 = \exp(\alpha_1 \frac{\overrightarrow{QP} \cdot \overrightarrow{QA} }{\| \overrightarrow{QA} \|^2} + \beta_1), \quad
	r_2 = \exp(\alpha_2 \frac{\overrightarrow{QP} \cdot \overrightarrow{QB} }{\| \overrightarrow{QB} \|^2} + \beta_2),
	\label{eq:rigidness}
\end{equation}
where $P$ is the projection of $X$ onto the connecting line; $\alpha_1$ and $\beta_1$ are learnable parameters to adjust the rigidness of bone $b_1$; $\alpha_2$ and $\beta_2$ adjust the rigidness of bone $b_2$. When the point $X$ moves closer to bone $b_1$, its rigidness about bone $b_1$ increases while its rigidness about bone $b_2$ decreases.

Based on the defined rigidness, we define the blending weights of a point with respect to bone $b_1$ and $b_2$ as
\begin{equation}
	w_1 = \frac{r_1}{r_1 + r_2}, \quad w_2 = \frac{r_2}{r_1 + r_2}.
	\label{eq:blending-weights}
\end{equation}
Given \cref{eq:blending-weights}, $w_1 + w_2=1$, which is crucial for perfect part seaming. As an explanation, when two parts undergo a relative rotation for angle $\theta$, their sections will have an angle gap of $\theta$. To maintain the connection, the sum of relative rotations $w_1 \theta + w_2 \theta$ must equals to $\theta$. Therefore, $w_1 + w_2=1$ is required.

\subsection{Optimization}
The complete supervision of our UNIF model is
\begin{equation}
	\mathcal{L}_{\text{}} = \mathcal{L}_{\text{recon}} 
	+ \lambda_{\text{unit}}\mathcal{L}_{\text{unit}} 
	+ \lambda_{\text{lim}}\mathcal{L}_{\text{lim}} 
	+ \lambda_{\text{sec}}\mathcal{L}_{\text{sec}} 
	+ \lambda_{\text{perim}}\mathcal{L}_{\text{perim}},
\end{equation}
where $\lambda_{\text{unit}}=0.1$, $\lambda_{\text{lim}}=1.0$, $\lambda_{\text{sec}}=0.01$, $\lambda_{\text{perim}}=0.001$.

We follow the same network architecture of IGR \cite{gropp2020igr} except that we lower the largest width of each MLP from 256 to 64 for a comparable number of parameters. 
For each frame of the raw scan sequence, we sample 5k points as surface points for the reconstruction term $\mathcal{L}_{\text{recon}}$; we also sample 5k points near the surface points with local disturbances ($\sigma_{\text{local}} \sim \mathcal{N}(0, 0.1)$) and another 5k points in the enlarged bounding box ($\sigma_{\text{global}}=1.5$) of the point cloud for the regularization terms $\mathcal{L}_{\text{unit}}$ and $\mathcal{L}_{\text{perim}}$.
We use an NVIDIA A40 GPU for each experiment with 4 scans in a batch. We train our model on each subject for 5k epochs using the Adam optimizer \cite{kingma2014adam} with a learning rate of 1e-3 and scale it down with a coefficient of 0.3 three times every 1k epochs. To extract surfaces from our learned neural implicit functions, we use the Marching Cubes algorithm \cite{lorensen1987marching} with the help of MISE \cite{mescheder2019occupancy} under a resolution of 256.

\section{Experiments}
\subsection{Settings}
\subsubsection{Datasets.}
We test our method on two datasets with raw scan sequences of clothed humans. The CAPE \cite{ma2020learning} dataset contains 15 subjects registered by SMPL \cite{loper2015smpl} with additional vertex offsets to model the clothes. Only four of the subjects have their raw scans released. Each of the four subjects has 4 to 6 sequences with 2 clothing types. We learn a model for each clothing type of a subject with one sequence left out for the extrapolation test. The length of each sequence ranges from about 200 to 550. For the training sequences, we use the first frame of every 10 frames for training and the fifth frame of every 10 frames for the interpolation test. The ClothSeq \cite{tiwari2021neural} dataset contains three subjects wearing loose clothes, therefore is more challenging. Each subject has one sequence, the length of which ranges from about 500 to 750. We use the first 80 percent frames with a stride of 10 for training, the last 20 percent frames for extrapolation test also with a stride of 10. For the interpolation test, we use frames from the first 80 percent with an offset of 5.

\subsubsection{Baselines.}
NASA \cite{deng2020nasa} is a typical part-based method that learns a group of occupancy networks anchored on joints. SCANimate \cite{saito2021scanimate} learns a forward and a backward skinning network with cycle consistency. SNARF \cite{chen2021snarf} conducts backward skinning with iterative root finding to improve generalizing to novel poses. 

\subsubsection{Metrics.}
For quantitative evaluation, we sample 100k points from each raw scan and our extracted surface, respectively. We report four metrics during our experiments including the point-to-surface distance (p2s), the recall rate, the Chamfer distance (CD), and the F-score. The point-to-surface distance is computed by the mean distance from a point in the raw scan to its closet point on our extracted surface. The recall rate counts the ratio of points with a point-to-surface distance lower than a threshold (1 mm). The Chamfer distance is the mean of the point-to-surface and the surface-to-point distance, and the F-score is the harmonic mean of recall and precision.

\begin{table}[t]
	\centering
	\tiny
	\renewcommand{\arraystretch}{1.0}
	\setlength{\tabcolsep}{1.4pt}
	\caption{Comparison with baselines on the CAPE \cite{ma2020learning} dataset. In the upper half rows are the results of the extrapolation test, which shows the generalization ability of a model, and the lower half are from the interpolation test, which shows the expressiveness of a model.}
	\label{tab:cape}
	
	\begin{tabular}{@{}ccrrrrrrrrrrrrrrrr@{}}
		\toprule
		\multirow{2}{*}{}                        & \multirow{2}{*}{seq.}         & \multicolumn{4}{c}{SCANimate}                                                                           & \multicolumn{4}{c}{SNARF}                                                                                       & \multicolumn{4}{c}{NASA}                                                                                & \multicolumn{4}{c}{Ours}                                                                               \\ \cmidrule(l){3-18} 
		&                               & \multicolumn{1}{c}{CD$\downarrow$} & \multicolumn{1}{c}{F1$\uparrow$} & \multicolumn{1}{c}{p2s$\downarrow$} & \multicolumn{1}{c|}{Rec.$\uparrow$} & \multicolumn{1}{c}{CD$\downarrow$} & \multicolumn{1}{c}{F1$\uparrow$} & \multicolumn{1}{c}{p2s$\downarrow$} & \multicolumn{1}{c|}{Rec.$\uparrow$} & \multicolumn{1}{c}{CD$\downarrow$} & \multicolumn{1}{c}{F1$\uparrow$} & \multicolumn{1}{c}{p2s$\downarrow$} & \multicolumn{1}{c|}{Rec.$\uparrow$} & \multicolumn{1}{c}{CD$\downarrow$} & \multicolumn{1}{c}{F1$\uparrow$} & \multicolumn{1}{c}{p2s$\downarrow$} & \multicolumn{1}{c}{Rec.$\uparrow$}  \\ \midrule
		\multicolumn{1}{c|}{\multirow{8}{*}{E}} & \multicolumn{1}{c|}{0032-SL} & 10.19                  & 66.16                  & 10.19                   & \multicolumn{1}{r|}{67.31}  & 10.71                  & 65.40                  & 10.68                   & \multicolumn{1}{r|}{66.96}          & 98.23                  & 15.78                  & 103.35                  & \multicolumn{1}{r|}{15.16}  & \textbf{8.06}          & \textbf{75.09}         & \textbf{7.87}           & \textbf{75.93}             \\
		\multicolumn{1}{c|}{}                    & \multicolumn{1}{c|}{0032-SS} & 9.89                   & 65.56                  & 9.45                    & \multicolumn{1}{r|}{66.54}  & 15.49                  & 49.58                  & 15.55                   & \multicolumn{1}{r|}{48.58}          & 131.79                 & 9.01                   & 74.52                   & \multicolumn{1}{r|}{11.81}  & \textbf{8.37}          & \textbf{72.55}         & \textbf{8.18}           & \textbf{72.86}             \\
		\multicolumn{1}{c|}{}                    & \multicolumn{1}{c|}{0096-SL} & 14.53                  & 56.97                  & 16.89                   & \multicolumn{1}{r|}{57.25}  & 12.19                  & 63.70                  & 14.35                   & \multicolumn{1}{r|}{63.93}          & 92.74                  & 10.69                  & 93.72                   & \multicolumn{1}{r|}{10.53}  & \textbf{10.40}         & \textbf{64.36}         & \textbf{10.04}          & \textbf{65.54}             \\
		\multicolumn{1}{c|}{}                    & \multicolumn{1}{c|}{0096-SS} & 11.25                  & 64.89                  & 11.51                   & \multicolumn{1}{r|}{65.50}  & 23.57                  & \textbf{72.47}         & 24.68                   & \multicolumn{1}{r|}{\textbf{73.42}} & 101.51                 & 14.37                  & 86.82                   & \multicolumn{1}{r|}{14.77}  & \textbf{8.74}          & 71.57                  & \textbf{8.50}           & 72.87                      \\
		\multicolumn{1}{c|}{}                    & \multicolumn{1}{c|}{0159-SL} & 7.93                   & 75.24                  & 7.49                    & \multicolumn{1}{r|}{76.96}  & 29.34                  & 68.65                  & 33.26                   & \multicolumn{1}{r|}{67.22}          & 118.10                 & 8.08                   & 153.04                  & \multicolumn{1}{r|}{7.47}   & \textbf{6.64}          & \textbf{82.42}         & \textbf{6.28}           & \textbf{83.05}             \\
		\multicolumn{1}{c|}{}                    & \multicolumn{1}{c|}{0159-SS} & 6.52                   & 84.34                  & 6.15                    & \multicolumn{1}{r|}{85.71}  & 20.76                  & 78.39                  & 26.82                   & \multicolumn{1}{r|}{77.42}          & 85.46                  & 11.73                  & 81.09                   & \multicolumn{1}{r|}{12.37}  & \textbf{5.91}          & \textbf{86.20}         & \textbf{5.66}           & \textbf{87.61}             \\
		\multicolumn{1}{c|}{}                    & \multicolumn{1}{c|}{3223-SL} & 8.12                   & 77.95                  & 8.60                    & \multicolumn{1}{r|}{78.28}  & 25.29                  & 68.29                  & 30.17                   & \multicolumn{1}{r|}{67.20}          & 66.91                  & 21.31                  & 73.49                   & \multicolumn{1}{r|}{20.06}  & \textbf{6.24}          & \textbf{86.77}         & \textbf{5.47}           & \textbf{88.99}             \\
		\multicolumn{1}{c|}{}                    & \multicolumn{1}{c|}{3223-SS} & 9.45                   & 75.08                  & 10.93                   & \multicolumn{1}{r|}{74.24}  & 13.90                  & 83.83                  & 16.32                   & \multicolumn{1}{r|}{84.41}          & 70.15                  & 22.78                  & 67.47                   & \multicolumn{1}{r|}{23.04}  & \textbf{5.61}          & \textbf{87.88}         & \textbf{5.31}           & \textbf{89.61}             \\ \midrule
		\multicolumn{1}{c|}{\multirow{8}{*}{I}}  & \multicolumn{1}{c|}{0032-SL} & 6.86                   & 85.81                  & 6.80                    & \multicolumn{1}{r|}{88.76}  & 4.93                   & \textbf{95.51}         & 5.06                    & \multicolumn{1}{r|}{\textbf{97.93}} & 10.00                  & 74.16                  & 10.01                   & \multicolumn{1}{r|}{75.38}  & \textbf{4.14}          & 95.46                  & \textbf{3.72}           & 97.60                      \\
		\multicolumn{1}{c|}{}                    & \multicolumn{1}{c|}{0032-SS} & 5.70                   & 90.45                  & 5.23                    & \multicolumn{1}{r|}{93.39}  & \textbf{4.07}          & \textbf{96.79}         & 3.99                    & \multicolumn{1}{r|}{\textbf{98.23}} & 10.28                  & 68.45                  & 10.38                   & \multicolumn{1}{r|}{69.26}  & 4.17                   & 95.30                  & \textbf{3.83}           & 97.01                      \\
		\multicolumn{1}{c|}{}                    & \multicolumn{1}{c|}{0096-SL} & 8.48                   & 89.50                  & 10.69                   & \multicolumn{1}{r|}{91.94}  & 6.48                   & \textbf{96.93}         & 8.92                    & \multicolumn{1}{r|}{98.07}          & 15.47                  & 61.22                  & 18.34                   & \multicolumn{1}{r|}{62.08}  & \textbf{4.69}          & 96.05                  & \textbf{4.47}           & \textbf{98.33}             \\
		\multicolumn{1}{c|}{}                    & \multicolumn{1}{c|}{0096-SS} & 7.08                   & 82.76                  & 6.47                    & \multicolumn{1}{r|}{85.05}  & 4.05                   & 96.41                  & 3.84                    & \multicolumn{1}{r|}{97.84}          & 12.73                  & 67.40                  & 11.29                   & \multicolumn{1}{r|}{69.12}  & \textbf{3.74}          & \textbf{97.08}         & \textbf{3.42}           & \textbf{98.94}             \\
		\multicolumn{1}{c|}{}                    & \multicolumn{1}{c|}{0159-SL} & 5.18                   & 91.79                  & 4.35                    & \multicolumn{1}{r|}{96.01}  & 3.77                   & 96.35                  & 3.18                    & \multicolumn{1}{r|}{99.27}          & 11.82                  & 66.37                  & 11.39                   & \multicolumn{1}{r|}{69.81}  & \textbf{3.39}          & \textbf{96.80}         & \textbf{2.72}           & \textbf{99.91}             \\
		\multicolumn{1}{c|}{}                    & \multicolumn{1}{c|}{0159-SS} & 4.77                   & 93.75                  & 4.20                    & \multicolumn{1}{r|}{96.71}  & 3.42                   & 97.74                  & 3.18                    & \multicolumn{1}{r|}{99.19}          & 12.28                  & 65.86                  & 12.04                   & \multicolumn{1}{r|}{67.05}  & \textbf{2.94}          & \textbf{98.00}         & \textbf{2.69}           & \textbf{99.81}             \\
		\multicolumn{1}{c|}{}                    & \multicolumn{1}{c|}{3223-SL} & 5.31                   & 93.40                  & 5.26                    & \multicolumn{1}{r|}{96.81}  & 5.06                   & 95.70                  & 5.86                    & \multicolumn{1}{r|}{95.84}          & 8.17                   & 84.47                  & 7.92                    & \multicolumn{1}{r|}{85.61}  & \textbf{3.89}          & \textbf{96.55}         & \textbf{3.07}           & \textbf{99.58}             \\
		\multicolumn{1}{c|}{}                    & \multicolumn{1}{c|}{3223-SS} & 4.89                   & 94.09                  & 4.74                    & \multicolumn{1}{r|}{97.15}  & 3.76                   & 97.68                  & 3.88                    & \multicolumn{1}{r|}{99.24}          & 7.80                   & 86.61                  & 6.95                    & \multicolumn{1}{r|}{87.93}  & \textbf{3.09}          & \textbf{97.81}         & \textbf{2.84}           & \textbf{99.68}             \\ \bottomrule
	\end{tabular}
\end{table}

\begin{table}[t]
	\centering
	\tiny
	\renewcommand{\arraystretch}{1.0}
	\setlength{\tabcolsep}{1.8pt}
	\caption{Comparison with baselines on the ClothSeq \cite{tiwari2021neural} dataset. In the upper half rows are the results of the extrapolation test, and the lower half are from the interpolation test.}
	\label{tab:clothseq}
	
	\begin{tabular}{@{}ccrrrrrrrrrrrrrrrr@{}}
		\toprule
		\multirow{2}{*}{}                       & \multirow{2}{*}{seq.}   & \multicolumn{4}{c}{SCANimate}                                                                           & \multicolumn{4}{c}{SNARF}                                                                               & \multicolumn{4}{c}{NASA}                                                                                & \multicolumn{4}{c}{Ours}                                                                               \\ \cmidrule(l){3-18} 
		&                               & \multicolumn{1}{c}{CD$\downarrow$} & \multicolumn{1}{c}{F1$\uparrow$} & \multicolumn{1}{c}{p2s$\downarrow$} & \multicolumn{1}{c|}{Rec.$\uparrow$} & \multicolumn{1}{c}{CD$\downarrow$} & \multicolumn{1}{c}{F1$\uparrow$} & \multicolumn{1}{c}{p2s$\downarrow$} & \multicolumn{1}{c|}{Rec.$\uparrow$} & \multicolumn{1}{c}{CD$\downarrow$} & \multicolumn{1}{c}{F1$\uparrow$} & \multicolumn{1}{c}{p2s$\downarrow$} & \multicolumn{1}{c|}{Rec.$\uparrow$} & \multicolumn{1}{c}{CD$\downarrow$} & \multicolumn{1}{c}{F1$\uparrow$} & \multicolumn{1}{c}{p2s$\downarrow$} & \multicolumn{1}{c}{Rec.$\uparrow$}  \\ \midrule
		\multicolumn{1}{c|}{\multirow{3}{*}{E}} & \multicolumn{1}{c|}{JP} & 14.33                  & 56.25                  & 14.29                   & \multicolumn{1}{r|}{58.02}  & 17.72                  & 58.49                  & 21.58                   & \multicolumn{1}{r|}{59.16}  & 69.76                  & 16.88                  & 68.24                   & \multicolumn{1}{r|}{16.59}  & \textbf{13.04}         & \textbf{58.60}         & \textbf{11.24}          & \textbf{62.06}             \\
		\multicolumn{1}{c|}{}                   & \multicolumn{1}{c|}{JS} & \textbf{11.05}         & 61.26                  & 10.85                   & \multicolumn{1}{r|}{62.58}  & 13.40                  & 57.48                  & 13.91                   & \multicolumn{1}{r|}{57.72}  & 116.14                 & 8.51                   & 97.25                   & \multicolumn{1}{r|}{9.61}   & 11.84                  & \textbf{65.94}         & \textbf{9.00}           & \textbf{70.03}             \\
		\multicolumn{1}{c|}{}                   & \multicolumn{1}{c|}{SP} & 14.32                  & 54.06                  & 14.19                   & \multicolumn{1}{r|}{54.80}  & 15.06                  & 60.22                  & 16.47                   & \multicolumn{1}{r|}{60.19}  & 65.73                  & 19.93                  & 39.26                   & \multicolumn{1}{r|}{21.29}  & \textbf{12.10}         & \textbf{65.90}         & \textbf{9.81}           & \textbf{69.46}             \\ \midrule
		\multicolumn{1}{c|}{\multirow{3}{*}{I}} & \multicolumn{1}{c|}{JP} & 10.05                  & 71.78                  & 7.38                    & \multicolumn{1}{r|}{79.40}  & 8.43                   & 80.82                  & 8.67                    & \multicolumn{1}{r|}{83.92}  & 22.37                  & 44.95                  & 21.97                   & \multicolumn{1}{r|}{45.42}  & \textbf{7.87}          & \textbf{84.66}         & \textbf{5.47}           & \textbf{90.01}             \\
		\multicolumn{1}{c|}{}                   & \multicolumn{1}{c|}{JS} & 8.84                   & 74.84                  & 7.77                    & \multicolumn{1}{r|}{78.33}  & \textbf{8.81}          & 80.20                  & 7.86                    & \multicolumn{1}{r|}{81.80}  & 33.66                  & 31.61                  & 34.35                   & \multicolumn{1}{r|}{31.34}  & 8.89                   & \textbf{81.48}         & \textbf{5.96}           & \textbf{86.72}             \\
		\multicolumn{1}{c|}{}                   & \multicolumn{1}{c|}{SP} & 13.20                  & 57.74                  & 12.52                   & \multicolumn{1}{r|}{59.28}  & 11.21                  & 73.04                  & 11.08                   & \multicolumn{1}{r|}{74.40}  & 48.69                  & 38.02                  & 33.54                   & \multicolumn{1}{r|}{40.06}  & \textbf{10.18}         & \textbf{75.49}         & \textbf{7.42}           & \textbf{80.31}             \\ \bottomrule
	\end{tabular}
\end{table}

\subsection{Comparisons}
\begin{figure}[t]
	\centering
	\includegraphics[width=1.0\linewidth]{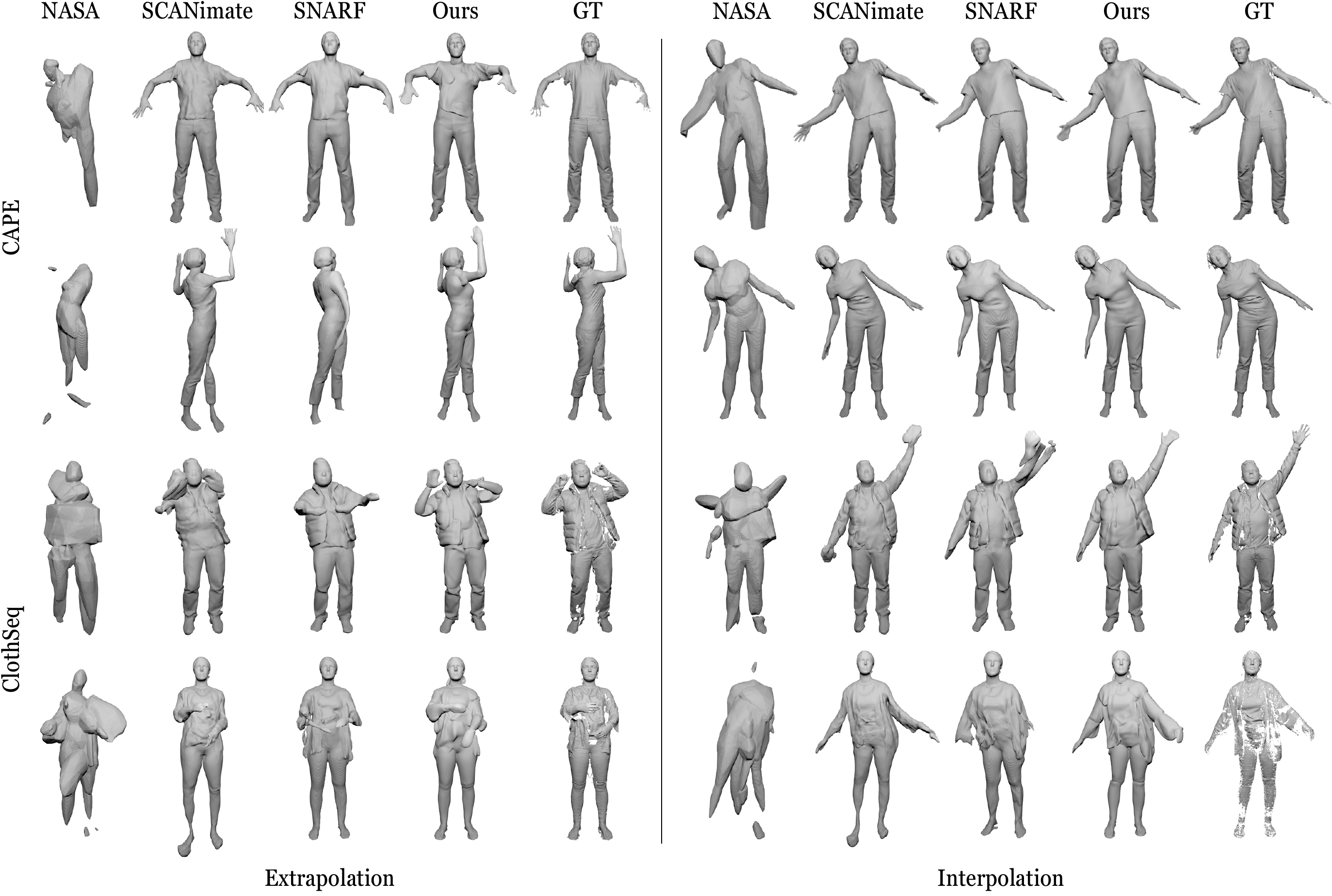}
	\caption{Qualitative comparison with baselines.}
	\label{fig:comparisons}
\end{figure}

We show quantitative results in \cref{tab:cape} and \cref{tab:clothseq} and qualitative results in \cref{fig:comparisons}. Our method shows clear superiority over NASA \cite{deng2020nasa} (also a part-based method) and outperforms SCANimate \cite{saito2021scanimate} and SNARF \cite{chen2021snarf} in most cases.

The extrapolation test is extremely challenging, especially on CAPE \cite{ma2020learning} because the test poses differ a lot from the limited training poses. Therefore, NASA \cite{deng2020nasa} completely collapses; SCANimate \cite{saito2021scanimate} and SNARF \cite{chen2021snarf} produce distortions due to bad neural skinning weights. Our method shows higher robustness under novel poses, benefiting from the proper part division and the generalizable non-rigid deformation modeling.

In the interpolation test, NASA \cite{deng2020nasa} exhibits reasonable results on CAPE \cite{ma2020learning} but has difficulty in partitioning and reconstructing the subjects in ClothSeq \cite{tiwari2021neural}. Our method produces visually comparable results with SCANimate \cite{saito2021scanimate} and SNARF \cite{chen2021snarf} on CAPE \cite{ma2020learning}. However, on the ClothSeq \cite{tiwari2021neural} dataset, where clothes are much more complex, our method makes less body distortion or extra surfaces and reconstructs the pose of the subjects more precisely.

\subsection{Visualization and Analysis}

\subsubsection{Ablation Study.}

\begin{figure}[t]
	\centering
	\includegraphics[width=0.95\linewidth]{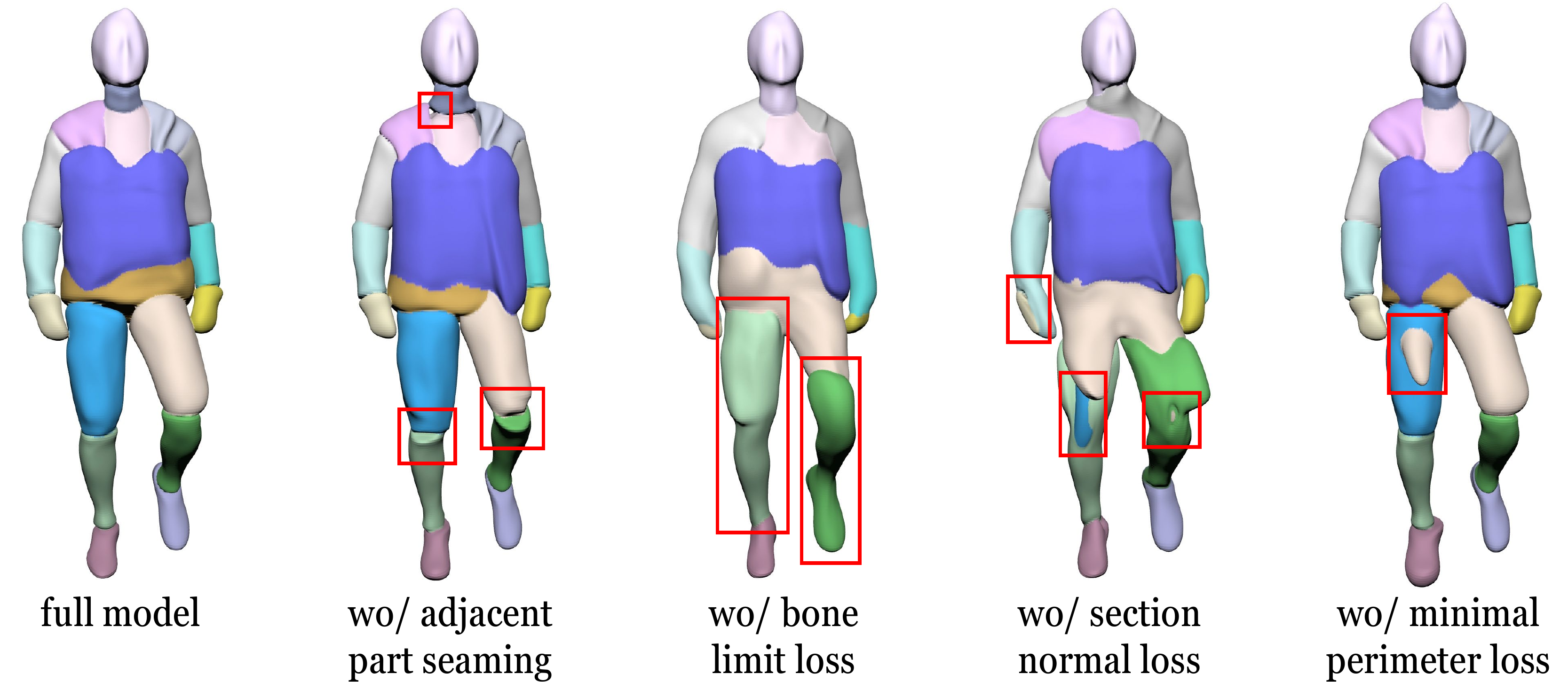}
	\caption{Ablation study on main components of our method.}
	\label{fig:ablation}
\end{figure}

To validate the effectiveness of our main components, we run experiments with each of them disabled and visualize parts in an unseen pose at the early stage of training in \cref{fig:ablation}. Compared to our full model, dropping the adjacent part seaming algorithm leaves the model almost rigid (\eg, the sections near the knees are exposed and the neck is not completely connected to the body). When disabling the bone limit loss, we lose the restriction on the boundary of a part. Then we see the left foot and leg of the man falsely included in the same part. However, merely using the bone limit loss is not sufficient. If we drop the section normal loss, the model converges to a bad partition where the surface does go across the joint but the main body of the part lies somewhere else. Finally, the minimal perimeter loss is also necessary to suppress extra surfaces such as the one on the right leg.

\begin{figure}[t]
	\centering
	\includegraphics[width=0.95\linewidth]{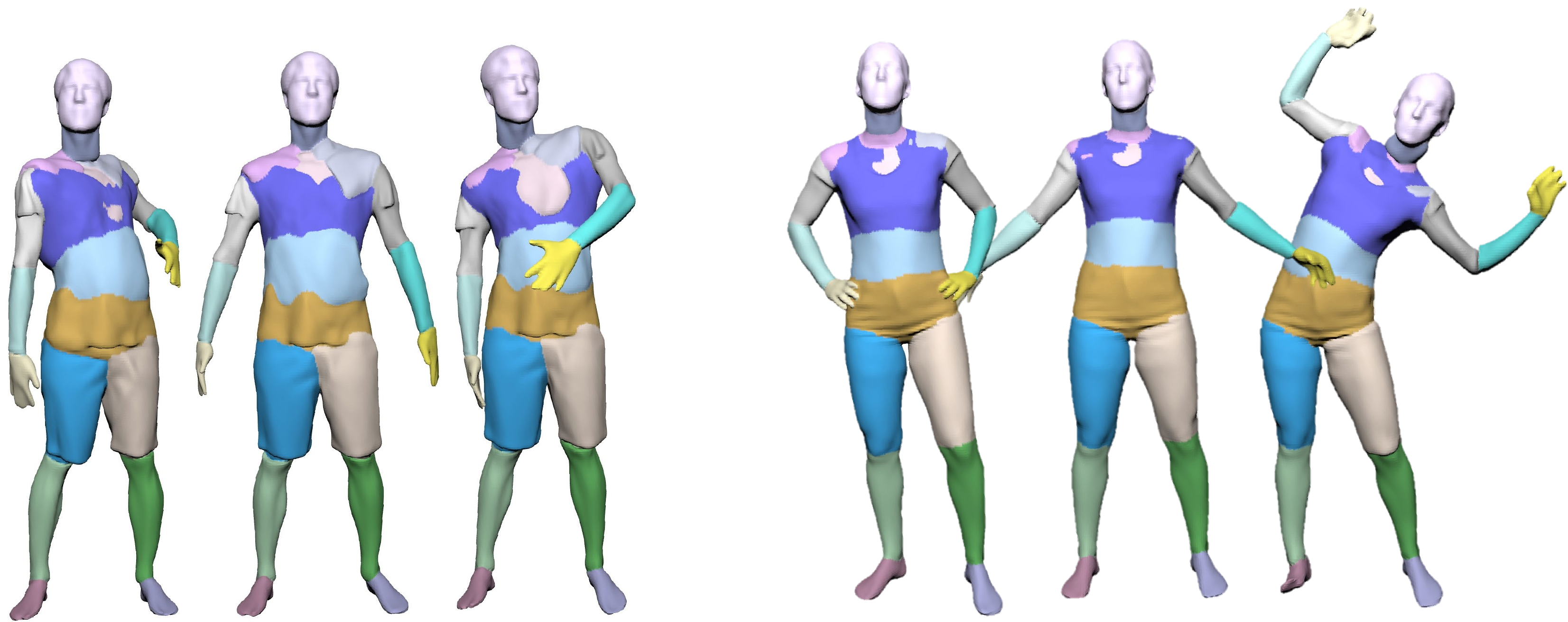}
	\caption{Visualization of learned parts during animation.}
	\label{fig:part-consistency}
\end{figure}

\subsubsection{Limitations.}
Since the \#13 and \#14 joints of SMPL \cite{loper2015smpl} are too close to the spine, our method learns a small chest and large shoulders. When shoulders move drastically, our model converges to overlapped parts to reconstruct the shape. Therefore, we can observe inconsistent part division around the chest during animation (\cref{fig:part-consistency}). The current framework does not model the dynamics of loose clothes. Seams between parts are still visible due to the generalization problem caused by the pose condition vector.


\section{Conclusions}
We present a novel method for clothed human reconstruction and animation. We explore initialization and regularization strategies to learn body parts without ground-truth part labels. Towards a higher generalization ability to novel poses, we propose an adjacent part seaming algorithm to model non-rigid deformations by explicitly modeling the interaction between parts. Experiments on two datasets validate the effectiveness of our method.

\noindent\textbf{Acknowledgments:}
The work is supported by National Key R\&D Program of China (2018AAA0100704), 
NSFC \#61932020, \#62172279, Science and Technology Commission of Shanghai Municipality (Grant No. 20ZR1436000), and  "huguang Program" supported by Shanghai Education Development Foundation and Shanghai Municipal Education Commission. 
This work is supported by NTU NAP, MOE AcRF Tier 2 (T2EP20221-0033), and under the RIE2020 Industry Alignment Fund – Industry Collaboration Projects (IAF-ICP) Funding Initiative, as well as cash and in-kind contribution from the industry partner(s).

\clearpage
\appendix

\section{Network Architecture}
\begin{figure}[h]
    \centering
    \includegraphics[width=0.8\linewidth]{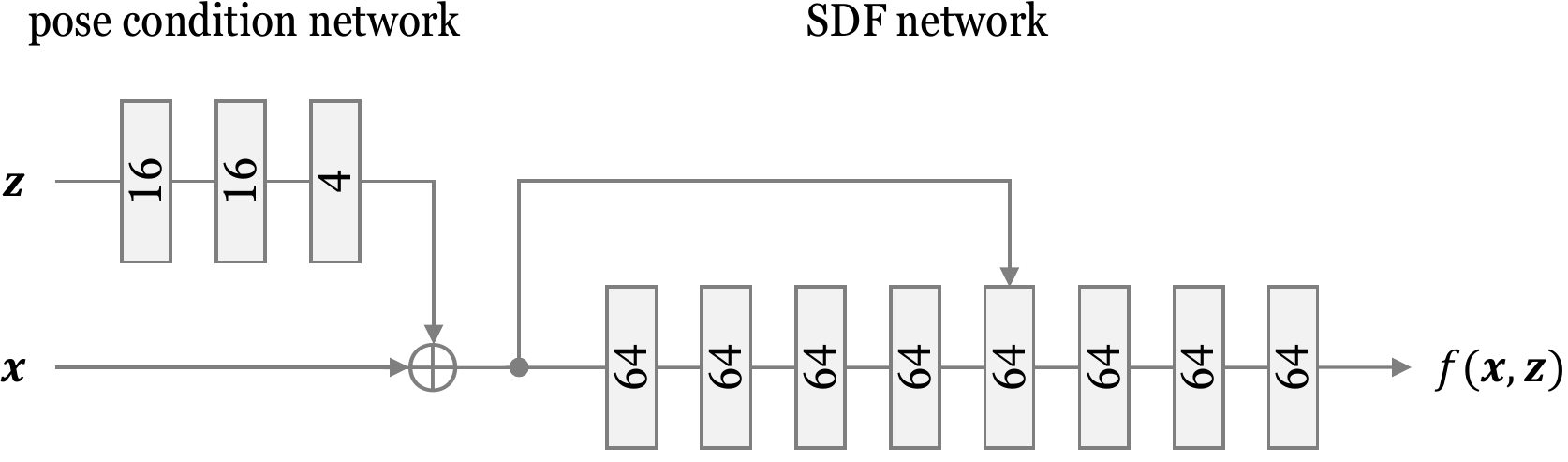}
    \caption{The architecture of each neural implicit function. Every gray box represents a linear layer with its output dimension marked on the box. Every linear layer except the last one is followed by an activation layer (Softplus \cite{zheng2015improving} with $\beta=100$).}
    \label{fig:arch}
\end{figure}

Our method uses multiple neural implicit functions to model a shape. All the networks share the same structure as shown in \cref{fig:arch}. The SDF networks are initialized with geometric initialization \cite{atzmon2020sal}, and the pose condition networks are initialized with Kaiming initialization \cite{he2015delving}. To relieve the conflict between the pose condition and geometric initialization \cite{atzmon2020sal}, we initialize the last layer of each pose condition network with weights sampled from the distribution $\mathcal{N}(0, 1e-5)$. We use weight normalization \cite{salimans2016weight} on SDF networks for stable training.

\section{The Union Operation}
During shape learning, taking the minimum means that supervising a data point only updates one network. This makes the model more sensitive to initialization and extreme body poses because once a body part is captured by a false network, it is hard for other networks to fight back due to the lack of chances to update. An alternative way is by using the smooth minimum function
\begin{equation}
    d = \frac{\sum_{n=1}^{N} d^{(n)} e^{-\beta d^{(n)}}}{\sum_{n=1}^{N} e^{-\beta d^{(n)}}},
\end{equation}
where $\beta$ controls the steepness of the weighting ratio. However, this function makes steeper blending for points farther away from the surface. Since we only care about the difference between $\min_{1\leq n \leq N}{d^{(n)}}$ and other outputs instead of their concrete values, we use the following function in practice:
\begin{equation}
    d = \min_{1\leq n \leq N}{d^{(n)}} + \frac{\sum_{n=1}^{N} \Delta d^{(n)} e^{-\beta \Delta d^{(n)}}}{\sum_{n=1}^{N} e^{-\beta \Delta d^{(n)}}},
\end{equation}
where $\Delta d^{(n)} = d^{(n)} - \min_{1\leq n \leq N}{d^{(n)}}$, and $\beta=200$.

\section{Minimal Perimeter Loss}
Our inspiration comes from PHASE \cite{lipman2021phase}, a method that bridges the signed distance field and the occupancy field with a logarithm transformation. To encourage learning a tight surface, PHASE \cite{lipman2021phase} applies a punishment on the norm of the gradient of the occupancy field while maintaining the unit-gradient-norm property in the SDF field. This works because the norm of the gradient for occupancy achieves the highest at the zero level and minimizing the norm of gradient is equivalent to suppressing all the surfaces. 

Since an SDF has a constant norm of gradient, we apply the sigmoid function ($\sigma(x) = \frac{1}{1 + e^{- \beta x}}$) on it to construct a peak of norm of gradient at the zero level. Then we can suppress surfaces by minimizing the norm of the gradient in the constructed field while maintaining the unit-gradient-norm property in the SDF field.

Our experiments also validate that this minimal perimeter loss is essential to suppress periodic extra surfaces when Fourier features \cite{tancik2020fourfeat} are used. However, since Fourier features can hurt the initialization of our method, we do not use it unless specified.

\section{Blending weights from ``Competing Bones''.}
\subsection{Derivation}
\begin{figure}
    \centering
    \begin{subfigure}[t]{0.32\textwidth}
        \centering
        \includegraphics[width=0.88\textwidth]{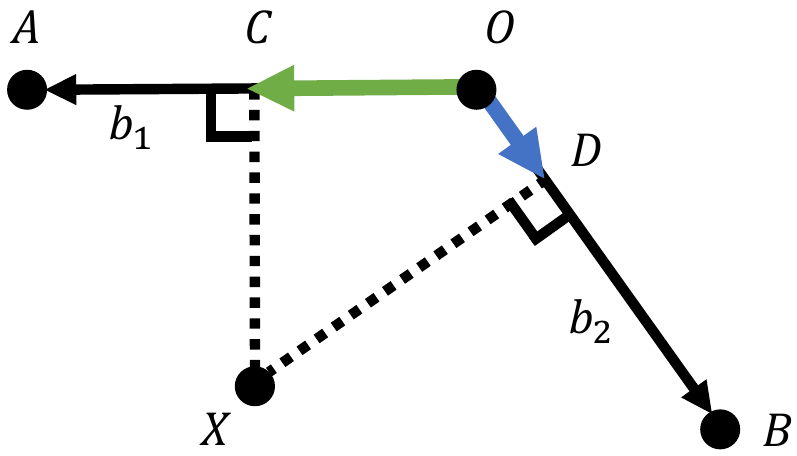}
        \caption{Rigidness defined by the projection of a point on bones with a large angle.}
        \label{fig:competing-bones_a}
    \end{subfigure}
    \hfill
    \begin{subfigure}[t]{0.33\textwidth}
        \centering
        \includegraphics[width=0.56\textwidth]{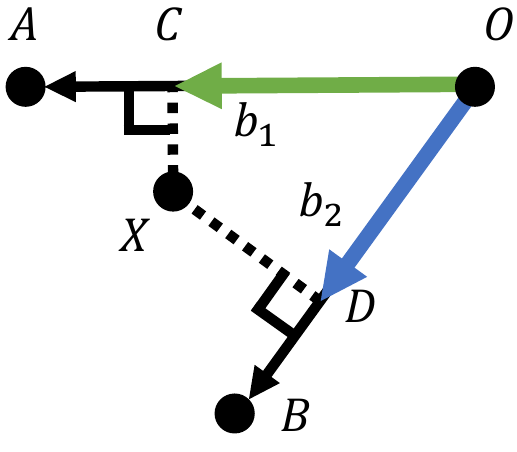}
        \caption{Rigidness defined by the projection of a point on bones with a small angle.}
        \label{fig:competing-bones_b}
    \end{subfigure}
    \hfill
    \begin{subfigure}[t]{0.32\textwidth}
        \centering
        \includegraphics[width=0.7\textwidth]{fig/competing-bones_c.pdf}
        \caption{Rigidness defined by the projection of a point on the connecting line.}
        \label{fig:competing-bones_c}
    \end{subfigure}
    \caption{2D toy examples to illustrate rigidness computation.}
\end{figure}

In the main text, we have defined the tendency of a point to stay static on a part as rigidness. Now we present the derivation of the concrete definition. Based on the common sense and the visualization of LBS weights of SMPL \cite{loper2015smpl}, we adopt a prior that the farther a point is away from the joint the more rigid it is. Therefore, we define a rigidness field by the projection of a point on the bone. For the example in \cref{fig:competing-bones_a}, the rigidness of bone $b_1$ and $b_2$ with respect to joint $O$ are defined as
\begin{equation}
    r_1 = \exp(\frac{\overrightarrow{OC} \cdot \overrightarrow{OA} }{\| \overrightarrow{OA} \|^2}), \quad
    r_2 = \exp(\frac{\overrightarrow{OD} \cdot \overrightarrow{OB} }{\| \overrightarrow{OB} \|^2}),
\end{equation}
For the point $X$ in \cref{fig:competing-bones_a}, when it moves closer to the farther end of bone $b_1$ than $b_2$, its rigidness on bone $b_1$ increases while its rigidness on bone $b_2$ decreases. However, when the angle between two linked bones is less than 90 degree (\cref{fig:competing-bones_b}), moving along a bone leads to increase of both bones, which is incorrect. Therefore, we correct the rigidness definition as shown in \cref{fig:competing-bones_c}. In the corrected version, we connect the end points of both bones and split it with point $Q$ by the ratio of bone lengths. Then we can compute the rigidness of point $X$ with respect to either bone by
\begin{equation}
    r_1 = \exp(\alpha_1 \frac{\overrightarrow{QP} \cdot \overrightarrow{QA} }{\| \overrightarrow{QA} \|^2} + \beta_1), \quad
    r_2 = \exp(\alpha_2 \frac{\overrightarrow{QP} \cdot \overrightarrow{QB} }{\| \overrightarrow{QB} \|^2} + \beta_2),
    \label{eq:rigidness}
\end{equation}
where P is the projection point of $X$ onto the connecting line; $\alpha_1$, $\alpha_2$, $\beta_1$, and $\beta_2$ are learnable parameters to adjust the rigidness of each bone. The corrected definition of rigidness works well regardless of the angle between the bones. Interestingly, when the angle between bones is 180 degree, the corrected version is equivalent to the first one.

\subsection{Rigidness Coefficients Learning}
\begin{figure}[t]
    \centering
    \includegraphics[width=1\linewidth]{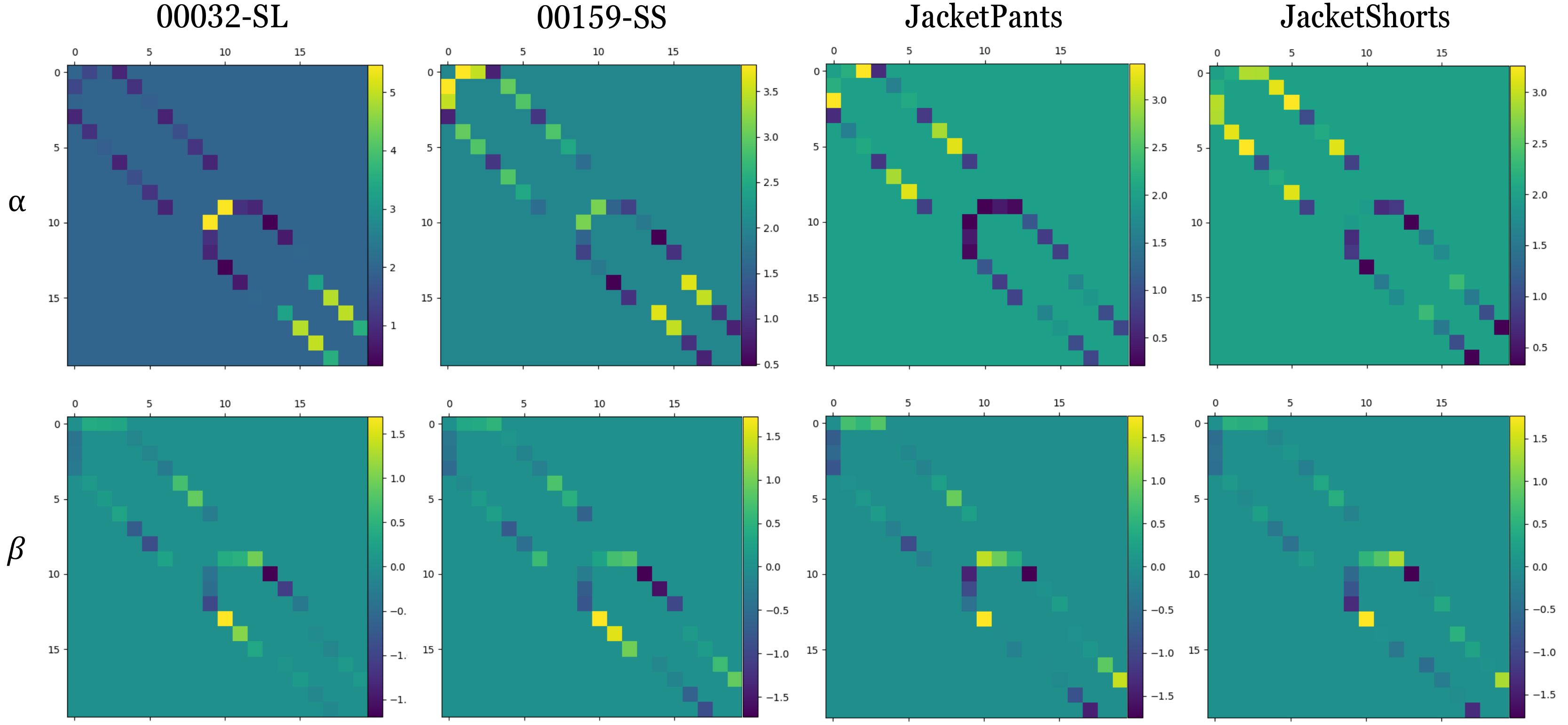}
    \caption{The learned rigidness coefficient matrices. The first row corresponds to the scaling factor $\alpha$, and the second row contains the bias factor $\beta$.}
    \label{fig:rigidness_param}
\end{figure}

Since the rigidness coefficients, the scaling factor $\alpha$ and the bias factor $\beta$, are defined for every pair of adjacent parts, we store them in matrices. Empirically, we init the matrices for $\alpha$ to 2 and the matrices for $\beta$ to 0. After the models have converged, we plot the rigidness coefficient matrices in \cref{fig:rigidness_param}. We observe that the matrices for the scaling factor $\alpha$ are symmetric, while the matrices for the bias factor $\beta$ are skew-symmetric. Note that only partial elements deviate from the initialization values because the connections between parts are sparse.

\section{Illustration of Adjacent Part Seaming}

\begin{figure}[h]
    \centering
    \includegraphics[width=0.8\linewidth]{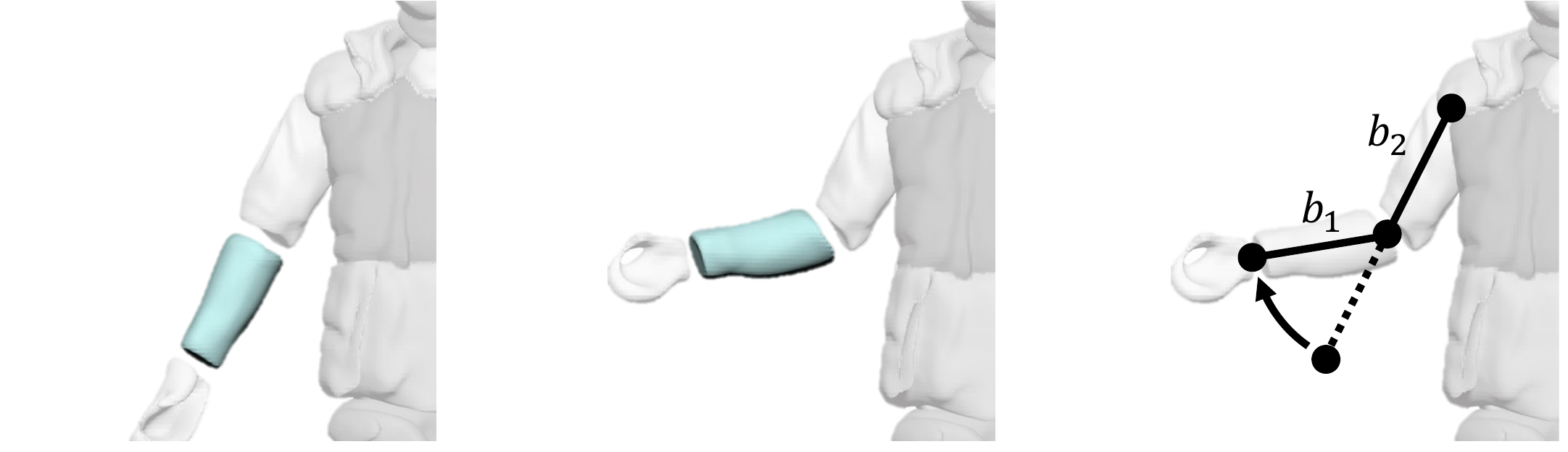}
    \caption{Non-rigid deformation produced by adjacent part seaming. When the bones $b_1$ and $b_2$ undergo a relative rotation, our adjacent part seaming algorithm guarantees the alignment of their sections. We disconnect adjacent parts to better visualize the sections and to confirm that parts are not overlapped.}
    \label{fig:aps_demo}
\end{figure}

\begin{figure}[h]
    \centering
    \includegraphics[width=1\linewidth]{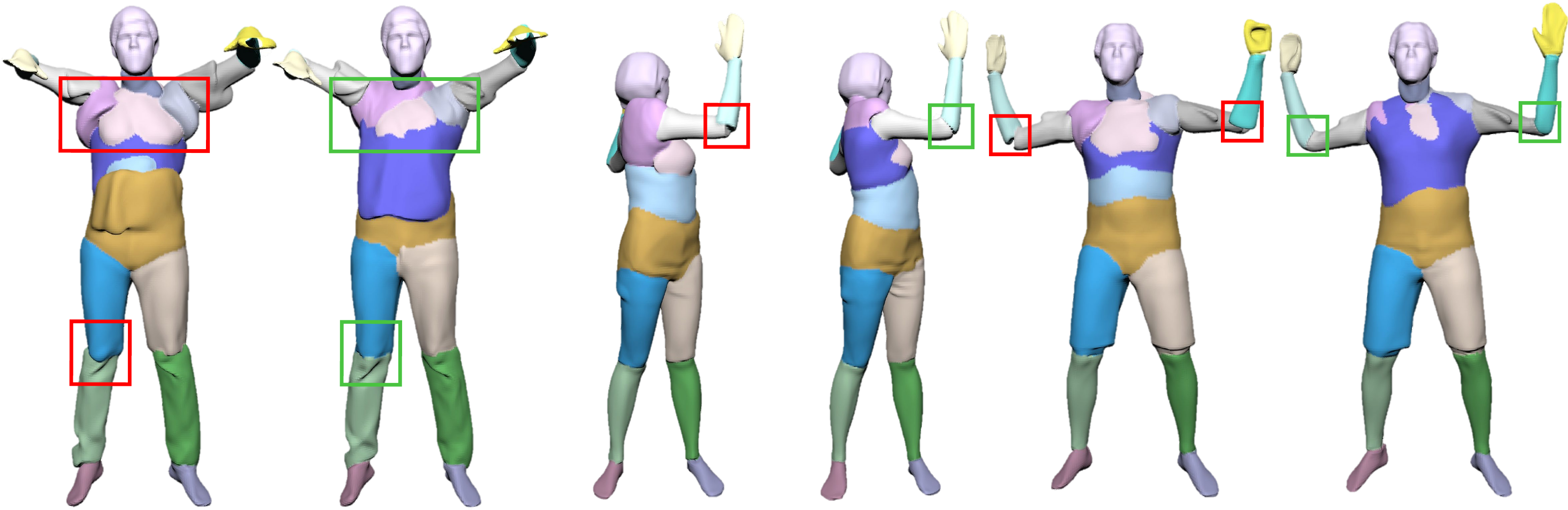}
    \caption{Results of our method under novel poses with (green boxes) and without (red boxes) adjacent part seaming.}
    \label{fig:ablation_aps}
\end{figure}

To illustrate the effect of adjacent part seaming (APS), we present a zoom-in example in \cref{fig:aps_demo}. When bone $b_1$ and bone $b_2$ undergo a relative rotation, the part on bone $b_1$ deforms so that its section across the joint can still align with the section of bone $b_2$.
Furthermore, we compare the results under novel poses with and without APS in \cref{fig:ablation_aps}. Disabling APS leaves the model almost rigid, resulting in artifacts like cracks at elbows and knees.

%
%
\bibliographystyle{splncs04}
\bibliography{egbib}

\end{document}